\documentclass{article}
\usepackage{spconf,amsmath,graphicx}

\usepackage{todonotes}

\title{Cascaded Models with Cyclic Feedback for Direct Speech Translation}
\name{Tsz Kin Lam$^{\star}$ \qquad Shigehiko Schamoni$^{\star, \dagger}$ \qquad Stefan Riezler$^{\star,\dagger}$}
\address{$^{\star}$Computational Linguistics \& $^\dagger$IWR, Heidelberg University, Germany \\
  {\tt \{lam,schamoni,riezler\}@cl.uni-heidelberg.de}
}
\begin{document}
%
\maketitle
%
%
\begin{abstract}
Direct speech translation describes a scenario where only speech inputs and corresponding translations are available. Such data are notoriously limited. We present a technique that allows cascades of automatic speech recognition (ASR) and machine translation (MT) to exploit in-domain direct speech translation data in addition to out-of-domain MT and ASR data.  After pre-training MT and ASR, we use a feedback cycle where the downstream performance of the MT system is used as a signal to improve the ASR system by self-training, and the MT component is fine-tuned on multiple ASR outputs, making it more tolerant towards spelling variations. A comparison to end-to-end speech translation using components of identical architecture and the same data shows gains of up to 3.8 BLEU points on LibriVoxDeEn and up to 5.1 BLEU points on CoVoST for German-to-English speech translation.
\end{abstract}
\section{Introduction}
\label{sec:intro}

Direct end-to-end automatic speech translation (AST) models \cite{WeissETAL:17} have been shown to overcome the error propagation issues of traditional cascades of automatic speech recognition (ASR) and machine translation (MT) \emph{if enough in-domain parallel data of source audio and text translation are available}\cite{SperberETAL:19}. 
Considerable research effort has thus been invested in improving the data efficiency of direct sequence-to-sequence AST, either by better exploitation of out-of-domain speech and translation resources in sophisticated information passing in multi-task approaches \cite{WeissETAL:17,SperberETAL:19,BerardETAL:18,AnastasopoulosChiang:18}, or by synthesizing parallel data by back-translation approaches \cite{JiaETAL:19,PinoETAL:19}. However, as recently shown by \cite{SperberPaulik:20}, problems like domain mismatch and error propagation might be re-introduced by exploiting out-of-domain data and by information-passing in end-to-end AST. On the other hand, cascaded models seem still to benefit more from out-of-domain data to directly improve their MT and ASR components, than end-to-end systems can exploit such data by multi-task learning \cite{SperberETAL:19,BerardETAL:18,PinoETAL:19}. 
One question in the ongoing competition between end-to-end and cascaded models is which paradigm is preferable in low-resource scenarios where only a few thousand parallel data of recorded speech and text translation, but no further in-domain data to train MT and ASR separately, are available. Such a scenario is real for endangered languages \cite{DuongETAL:16}, and it corresponds to the status quo in speech translation where copious amounts of training data are mostly available only in form of out-of-domain MT or ASR data.
In this paper, we focus on low-resource direct speech translation. We present an experiment that starts with small amounts of in-domain direct speech translation data (7k to 59k sentences), and varies out-of-domain data (from 15k to 150k for ASR, and from 50k to 2M  for MT).
We confirm common knowledge that end-to-end systems can better exploit in-domain direct speech translation data, while cascades outperform end-to-end systems \emph{if enough out-of-domain MT and ASR data} are available.

The main contribution of our paper is a novel adaptation cycle that allows ASR-MT cascades to also exploit direct speech translation data (that are useless for traditional cascades), and to eventually improve over end-to-end AST models by a wide margin.
The crucial ingredients of our cyclic feedback approach are firstly to train the MT system on $k$-best ASR outputs. This will teach the MT system to translate imperfect ASR outputs into correct foreign sentences. Furthermore, the evaluation performance of the produced MT output is used as  signal to improve the ASR system by selecting and weighting transcriptions leading to top-scoring translations as targets in self-training. This learning cycle will tune the ASR system towards producing transcriptions that perform well as translation inputs, thus improving the whole pipeline, without explicit parameter sharing or back-propagation. 
Our experiments on German-to-English speech translation on audio books  \cite{BeilharzETAL:20} and diverse domain \cite{WangETAL:20} corpora show that 3.8 or 5.1 BLEU points \cite{papineniETAL:02} can be gained over end-to-end systems on the respective datasets.

\section{Related Work}
\label{sec:related}

The problem of closing the domain gap between ASR output and text input to MT and has been addressed already in the framework of Statistical Machine Translation (SMT), by training SMT systems on automatically transcribed speech \cite{PeitzETAL:12}, or by augmenting SMT translation models with simulated acoustic confusions \cite{TsvetkovETAL:14}. 
In the area of neural sequence-to-sequence learning, similar approaches have been applied to ASR error correction, either directly by monolingual sequence-to-sequence transformation \cite{ManiETAL:20}, or by adapting the framework of generative adversarial networks to provide a language-model critic to improve ASR \cite{LiuETAL:19}.
Our work extends these ideas by using the performance improvement of downstream MT as learning signal in  self-training of ASR.

The idea of tuning ASR parameters for optimal downstream translation performance is even older, and has been applied successfully via minimum error rate training \cite{Och:03} of models that integrate ASR and MT features \cite{ZhangETAL:04,HeETAL:11}. In recent years, deep reinforcement learning has been the machine learning approach of choice in order to tune a neural sequence-to-sequence model to task-specific evaluation metrics \cite{KeneshlooETAL:19}. The crucial difference to our approach is that we reward the ASR system by an evaluation score that is grounded in a downstream application, and not by an evaluation metric that is task-specific to ASR. 
Furthermore, we do not rely on the machinery of reinforcement learning, but we follow a much simpler and more efficient training scheme where the downstream reward signal is used to select and weight ASR outputs for self-training.

Recently, out-of-domain pre-training has been combined with in-domain triplets for end-to-end fine-tuning \cite{LiuETAL19}. We use only speech-translation pairs to fine-tune our cascade. 
Speech-translation pairs for fine-tuning have also shown to be effective in a meta-learning scenario \cite{IndurthiETAL:20}. Their method was applied to larger datasets of 229k-275k pairs while our method works for very small datasets of 6.7k and 59k pairs. 


\begin{figure}[t!]
\centering
\includegraphics[width=.48\textwidth]{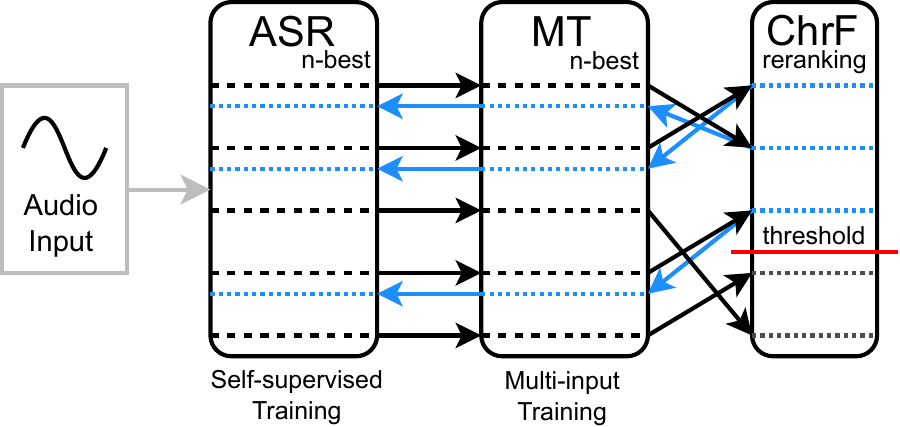}
\caption{Cyclic feedback in ASR-MT cascade.} 
\label{fig1}
\vspace{-3mm}
\end{figure}

\section{Cyclic Feedback in Direct Speech Translation}

Our cyclic feedback idea is based on self-training with a twist. The algorithm has two parts, where in one part the MT system is tuned to produce better translations for potentially noisy ASR outputs, and in the other part the ASR system is guided by the MT-output to generate transcriptions that led to higher scoring translations. Figure \ref{fig1} shows a cascaded model for a single German (\emph{de}) audio input. The model first produces $k$-best ASR transcriptions, which are fed as multiple inputs into an MT system. The translations are reranked according to their ChrF-score \cite{Popovic:15}, and a list of indices of translations exceeding a ChrF threshold (indicated by red dashed line) is kept. These indices are used as learning signal in a feedback cycle to select data to improve the ASR and MT components (indicated by blue dotted arrow). The MT system takes the selected data as input to learn how to translate imperfect ASR outputs into foreign reference sentences. The ASR system is trained in a self-supervised fashion using the selected data as reference transcriptions.

In our concrete implementation, at the beginning of the MT-adaptation loop, the ASR-system generates $k$-best ($k=8$) transcriptions via beam search. These outputs are then passed to the MT-system for translation. Based on the ChrF-score threshold of 0.4, we create a fine-tuning dataset by selecting corresponding ASR transcriptions as inputs for multi-input training. We additionally apply a weighting scheme that gives more weight to transcriptions that lead to better translations. The MT-system is then fine-tuned on this dataset via cross-entropy loss. This process is repeated until no further improvement on the dev set is obtained.
At the beginning of the ASR-adaptation loop, we again generate $k$-best ASR transcriptions and their translations (in the first loop, we can re-use the already generated transcriptions and translations from the last loop of MT-tuning). The ChrF-score of corresponding MT-outputs against the reference translations is used to indicate a dataset of audio data and the generated ASR transcriptions. A similar weighting scheme as for MT-adaptation is also applied, but as the ASR system is more sensible to errors, we increased the ChrF threshold to 0.6. The ASR-model is then fine-tuned with the generated data. This process can be repeated until the ASR training converges on the dev set. Finally, we start another cycle and enter the MT-adaptation loop until we observe no further improvement. 


\begin{table}[t]
\centering
\resizebox{.48\textwidth}{!}{%

\begin{tabular}{lccccc}
\hline
\textbf{Dataset} & \textbf{Purpose} & \multicolumn{3}{c}{\textbf{No. Sentences}} & \textbf{Avg. Words}  \\ 
 &  & 
 \multicolumn{1}{c}{\textbf{Train}} & \multicolumn{1}{c}{\textbf{Dev}} & \multicolumn{1}{c}{\textbf{Test}}  & \textbf{per Sentence} \\
\hline
Spoken Wikipedia & ASR PT & $160$k & $4$k & $4$k & $11.02$\\
WMT14 & MT PT & $2.1$M & $23$k & $22$k & $23.99$\\
LibriVoxDeEn & AST FT  & $6.7$k & $355$ & $1.1$k & $14.78$ \\
CoVoST De-En & AST FT & $59.0$k & $15.4$k & $145.5$k & $8.01$\\
\hline
\end{tabular}
}
\caption{Statistics of datasets used for pre-training (PT) and fine-tuning (FT) in our experiments.}
\label{tab:datainfo}
\vspace{-3mm}
\end{table}

\section{Experimental Setup}

\subsection{Datasets}
The first AST dataset that we use for fine-tuning on a new domain is LibriVoxDeEn \cite{BeilharzETAL:20}. This dataset consists of 86 classical German audio books with 547h of speech data. Of these 86 books, 19 books were published with German to English text alignments and can thus be used in AST training.
As we need very clean data in our experiments, we applied strong filtering on the LibriVoxEnDe data and selected only four out of the 19 audio books for fine tuning. 
For each book, we analyzed 30 randomly sampled parallel sentences and annotated the pairs as ``perfectly aligned'', ``partly aligned'', or ``not aligned''. We then selected the books which had perfectly aligned sentences in 80\% or more of the cases. These are \textit{The Picture of Dorian Gray} by Oscar Wilde, \textit{Casanovas Heimfahrt} by Arthur Schnitzler, \textit{Die Verwandlung} by Franz Kafka, and \textit{Undine} by Friedrich de la Motte Fouqué. These audio books are from the early 20th century except for the last one. 

The other dataset we use is the German-English portion of the CoVoST dataset \cite{WangETAL:20}. This dataset is built upon Common Voice \cite{ardila-etal-2020-common}. For each transcription-translation pair, there are multiple recordings for each transcription spoken by different speakers. 
Throughout our experiments on CoVoST we use the original data splits so our results are comparable to the baselines reported by \cite{WangETAL:20}. For efficiency, we reduced the dev set size to 15,351 examples by downsampling by $1/5$. 

For pre-training of ASR and MT components in an additional experiment, we employed large scale datasets from both areas. For ASR pre-training, we use the German Spoken Wikipedia (2.0) \cite{baumann2019spoken} corpus, a dataset containing more than 250h of aligned German sentences. For MT pre-training, we concatenated data from the official WMT14 English-German parallel data, namely Europarl v9, News Commentary v14, and IWSLT 2014. Table \ref{tab:datainfo} summarizes the statistics of each dataset and its splits.\footnote{Our LibriVoxDeEn data split: www.cl.uni-heidelberg.de/librivoxdeen/}

\begin{table}[t]
\centering
\resizebox{.48\textwidth}{!}{%
\begin{tabular}{lrrcccc}
\hline
\multicolumn{1}{c}{\textbf{Target}} & \multicolumn{2}{c}{\textbf{Fine-tuning data}} & \textbf{End-to-end} & \textbf{Untuned} & \textbf{Cascade} & \textbf{w/o ASR-}\\
\multicolumn{1}{c}{\textbf{domain}} & \multicolumn{1}{c}{\textbf{ASR}} & \multicolumn{1}{c}{\textbf{MT}} & \textbf{AST} & \textbf{cascade} &  & \textbf{tuning}\\
\hline
LibriVoxDeEn & $6.7$k & $6.7$k & $9.1$ & $8.1$ & $12.9 $ & $ 11.0$ \\
CoVoST De-En & $59.0$k & $59.0$k & $7.8$ & $11.3$ & $12.9 $ & $12.4 $ \\
\hline
\end{tabular}
}
\caption{Best results for end-to-end and cascade approaches.
}
\label{tab:baselines}
\vspace{-3mm}
\end{table}

\subsection{Systems}
\paragraph*{Speech Recognition and End-To-End Speech Translation}
We use a variant of a joint CTC-attention end-to-end framework \cite{LiuETAL:19}. The encoder consists of a VGG network followed by 3 blocks of BiLSTM layers. The VGG network reduces the temporal resolutions by a factor of 4. In each block, there is a BiLSTM layer with 256 hidden units per direction followed by a feed forward neural network with ELU activation \cite{ClevertETAL:16}. The decoder has one LSTM layer of size 256 which is connected to the encoder via location-based attention \cite{chorowski2015attention}. 
We applied Locked Dropout \cite{merity2018regularizing} to each BiLSTM block with a value of 0.2 and Embedding Dropout of 0.1 \cite{gal2016theoretically}.
The encoder takes 40-dimensional log Mel filter bank features with z-score normalization as input. Data instances containing audio longer than 2,000 frames were excluded.

For the speech recognition system, the German textual data was lowercased, all punctuation removed, and numbers were normalized to their spoken form using pre-processing tools from the \texttt{marytts}\footnote{https://github.com/marytts/marytts/} toolkit. 

\paragraph*{Machine Translation}
In order to better compare to the direct end-to-end AST system and to share pre-trained components, we use an LSTM-based architecture instead of a more sophisticated model such as the Transformer architecture \cite{vaswani2017attention}. Our encoder has 3 BiLSTM layers with 256 units per direction. The decoder consists of a single LSTM layer of size 256. Similar to the VGG encoder, we applied Locked Dropout with a value of 0.1 to the BiLSTM encoder. Dropout of 0.1 was applied to the target embedding. In addition, we shared parameters of the embeddings and the output layer.
Since both cascaded system and end-to-end AST system share the same data, we used 
a universal vocabulary of 10,000 subword units created with sentencepiece \cite{kudo2018sentencepiece} for all tasks.

\paragraph*{Training of End-To-End Speech Translation} In our low-resource scenario we assume that there are no audio-transcription-translation triplets in the target domain. 
This makes multi-task learning \cite{BerardETAL:18} on in-domain data impossible. 
We instead used a transfer-learning based method called the adapter \cite{bahar2019comparative}, which connects the pre-trained ASR encoder with the pre-trained MT decoder in a separate learned layer. 


\section{Experimental Results}

\paragraph*{Baselines and Best Results}

Table \ref{tab:baselines} gives an overview of our results compared to baselines. 
\cite{WangETAL:20} report 7.6 BLEU points for the end-to-end model on CovoST German-English data. Our end-to-end model performs at 7.8 BLEU, despite differences in experimental conditions such as character-level vs. sub-word units, different pre-training data, and our smaller decoder.
The results for our out-of-domain pre-trained cascade is at 11.3 BLEU for CoVoST, and improved to 12.9 BLEU by cyclic feedback.

To date no external baselines are available for a comparison on LibriVoxDeEn. We show improvements of 3.8 BLEU over our own end-to-end system, and of 4.8 BLEU over the untuned cascade.

\begin{table}[t!]
\centering
\resizebox{.48\textwidth}{!}{%
\begin{tabular}{lrrcccc}
\hline
\multicolumn{1}{c}{\textbf{Experiment}} & \multicolumn{2}{c}{\textbf{Pre-training data}} & \textbf{End-to-end} & \textbf{Untuned} & \textbf{+cyclic} \\
\multicolumn{1}{c}{\textbf{name}} & \multicolumn{1}{c}{\textbf{ASR}} & \multicolumn{1}{c}{\textbf{MT}} & \textbf{AST} & \textbf{cascade}  & \textbf{feedback}  \\
\hline
 librivox-100-100 & $159.5$k & $2.1$M & $9.1$ & $8.1$ & $12.9 $ \\
 librivox-100-10 & $159.5$k & $213.4$k & $8.8$ & $6.2$ & $9.4 $   \\
 librivox-100-2.5 & $159.5$k & $53.3$k & $8.8$ & $3.9$ &$5.4$  \\
 librivox-25-100 & $39.9$k & $2.1$M & $8.8$ & $6.2$ & $9.8$   \\
 librivox-25-10 & $39.9$k & $213.4$k & $8.9$ & $5.2$ & $7.5$   \\
 librivox-25-2.5 & $39.9$k & $53.3$k & $8.1$ & $3.2$ & $5.4$   \\
 librivox-10-100 & $15.9$k & $2.1$M & $8.4$ & $4.9$ & $6.8$   \\
 librivox-10-10 & $15.9$k & $213.4$k & $8.3$ & $4.1$ & $5.4 $   \\
 librivox-10-2.5 & $15.9$k & $53.3$k & $7.1$ & $2.6$ & $4.6 $   \\
\hline
\end{tabular}
}
\caption{Results on four audio books from LibriVoxDeEn under different data sizes for pre-training. 
}
\label{tab:results_librivox}
\vspace{-3mm}
\end{table}


\begin{table}[t!]
\centering
\resizebox{.48\textwidth}{!}{%

\begin{tabular}{lrrcccc}
\hline
\multicolumn{1}{c}{\textbf{Experiment}} & \multicolumn{2}{c}{\textbf{Pre-training data}} & \textbf{End-to-end} & \textbf{Untuned} & \textbf{+cyclic} & \\
\multicolumn{1}{c}{\textbf{name}} & \multicolumn{1}{c}{\textbf{ASR}} & \multicolumn{1}{c}{\textbf{MT}} & \textbf{AST} & \textbf{cascade} & \textbf{feedback} & \\
\hline
covost-100-100 & $159.5$k & $2.1$M & $7.8$ & $11.3$ & $12.9$ \\
covost-100-10 & $159.5$k & $213.4$k & $7.5$ & $9.4$ & $10.6$ \\
covost-100-2.5 & $159.5$k & $53.3$k & $7.0$ & $6.7$ & $7.6$ \\
covost-25-100 & $39.9$k & $2.1$M & $7.0$ & $7.8$ & $9.1 $ \\
covost-25-10 & $39.9$k & $213.4$k & $6.8$ & $6.6$ & $7.6$ \\
covost-25-2.5 & $39.9$k & $53.3$k & $6.5$ & $4.9$ & $5.7$ \\
covost-10-100 & $15.9$k & $2.1$M & $6.5$ & $5.2$ & $6.1$\\
covost-10-10 & $15.9$k & $213.4$k & $6.3$ & $4.5$  & $5.2$\\
covost-10-2.5 & $15.9$k & $53.3$k & $5.9$ & $3.5$  & $4.2$\\
\hline
\end{tabular}
}
\caption{Results on the German-English part of the CoVoST dataset under different data sizes for pre-training. 
}
\label{tab:results_covost_original_split}
\vspace{-3mm}
\end{table}


\paragraph*{LibriVoxDeEn} Table \ref{tab:results_librivox} shows the performance of end-to-end AST and cascaded system fine-tuned on LibriVoxDeEn under different sizes of pre-training data. The untuned cascaded system is 1 BLEU point behind the fine-tuned end-to-end AST system if both the ASR and MT components have access to all available data.
The end-to-end system is much less sensitive to the amounts of pre-training data: while the untuned cascade quickly drops in performance if both ASR and MT data are reduced, the end-to-end system suffers a reduction of at most 2 BLEU points. 
As soon as we add \emph{cyclic feedback}, the results of the cascaded system improve significantly and we see gains of up to 4.8 and 3.8 BLEU points over the untuned cascaded and the end-to-end AST system, respectively. The cyclic feedback approach is always better than the end-to-end system if the amount of pre-training data exceeds 100k for both ASR and MT.


\paragraph*{CoVoST De-En} 
Table \ref{tab:results_covost_original_split} shows that the untuned cascaded system matches or surpasses the end-to-end AST system if both the MT components have access to all available data. When using all pre-training data, the untuned cascaded system lies 3.5 BLEU points above the end-to-end system. We observe the following trend when the pre-training data is reduced: the end-to-end system has the largest drop when ASR-data is reduced, while the cascaded system is more sensitive to the quality of the MT-system, which is where the feedback signal for fine-tuning comes from. With \emph{cyclic feedback}, we again observe considerable improvements of up to 1.6 BLEU points for fine-tuning, but all in all lower than in the previous experiment. 
We assume that the simple textual structure of the CoVoST dataset is one main reason for the good performance of the untuned systems, and that the low number of unique examples in CoVoST's multi-speaker data provides too little variation for larger gains by cyclic feedback.


\paragraph*{Ablation Study} 
We also evaluated the contribution of ASR tuning based on feedback from the MT model to the cyclic feedback method. The results are listed in the last column of Table \ref{tab:baselines} titled ``without ASR-tuning'', where we applied fine-tuning only on the MT-part until no further improvement was observed. We get considerable improvements of 1.9 and 0.5 BLEU points on LibriVoxDeEn and Covost De-En, respectively. We also see improvements between 0.5 and 0.8 BLEU points in other cases with reduced data. This underlines the contribution of ASR tuning and shows the effectiveness of combining ASR and MT tuning in a cyclic feedback manner.

\begin{table}
\centering
\resizebox{.48\textwidth}{!}{%

\begin{tabular}{rll}
\hline
\# & & \textbf{Translation} \\
\hline
1 & Untuned & i wanted to say that i did not talk about cloabel.\\
 & Tuning & i wanted to say i don't talk to you..\\
 & Final & i wish i had not spoke of sibyl vane.\\
 & Reference & \textit{i wish now i had not told you about sibyl vane.}\\
\hline
2 & Untuned & i'm going to stay in the domain. he sadly said.\\
 & Tuning & i shall remain in the real dorian, he said. \\
 & Final & i shall stay with the real dorian, he said. \\
 & Reference & \textit{i shall stay with the real dorian, he said, sadly.}\\
\hline
3 &  Untuned & he had the life of his life, his life and his own resilience news of life. \\
 & Tuning & he had life in his life, his life, and his own vicious news of life. \\
 & Final & he life was determined to him, life and his own countless curiosity about life. \\
 & Reference & \textit{yes, life had decided that for him life, and his own infinite curiosity about life.}\\
\hline
4 &  Untuned & it is not only up to the bed in the bed-growth. \\
 & Tuning & i don't end up in the bed's misused, said gregor. \\
 & Final & it is only not to restraint in the bed, said gregor. \\ 
 & Reference & \textit{but i must not stay in bed uselessly, said gregor to himself.} \\
\hline

\end{tabular}

}
\caption{Four examples taken from our experiments on LibriVoxDeEn that illustrate the different steps in fine-tuning. 
}
\label{tab:examples}
\vspace{-3mm}
\end{table}

\paragraph*{Examples}

Table \ref{tab:examples} lists four examples from the LibriVox\-DeEn experiments to illustrate the process of fine-tuning via cyclic feedback. The \emph{untuned} translation comes from the untuned system and often contains misspelled words such as ``cloabel'' in Example 1 or ``bed-growth'' in Example 4. The \emph{tuning} translation is generated by the system after few steps of fine-tuning and already shows significant changes, e.g. ``domain'' is correctly transcribed as ``dorian'' in Example 2. In the same example we see that the almost correct sentence ``he sadly said'' is transformed to the subclause ``he said'', dropping the correct ``sadly'' adverb. 
The \emph{final} translation is the translation we receive from the final fine-tuned system. In Example 1, the nonsense word ``cloabel'' is correctly transformed to ``sibil vane''. 
Example 2 gives an almost perfect translation after the system is fine-tuned. 
The final examples in Examples 3 and 4 show typical examples in the test set, where the final translation is significantly better than the untuned version to the extend that the translation becomes understandable, but it is still far from perfect when compared to the \emph{reference}. 
Together with the modest BLEU scores we achieve overall, this again underlines the difficulty of the speech translation task on different domains.

\section{Conclusion}

We presented a novel domain adaptation technique for fine-tuning of cascaded AST models without the use of transcriptions. 
The key idea is to exploit translation quality as a signal to guide the ASR system to generate transcriptions that lead to better translations, and at the same time make the MT model more robust against transcription errors. In two low-resource domain adaptation scenarios on largely different domains we observe considerable gains over a comparable end-to-end system and the untuned out-of-domain cascaded system. 

\section{Acknowledgements}

This research was supported in part by the German research foundation (DFG) under grant RI-2221/4-1.

\bibliographystyle{IEEEbib}
{
\footnotesize
\bibliography{references}
}
%
%
\end{document}